\documentclass[final]{llncs}

\usepackage[arxiv]{eccv}
\usepackage{eccvabbrv}

\usepackage{graphicx}
\usepackage{booktabs}
\usepackage{multirow}
\usepackage[accsupp]{axessibility}  % Improves PDF readability for those with disabilities.

\usepackage{hyperref}

\usepackage{orcidlink}

\usepackage{cuted}
\def\model{LPH-VTON}

\begin{document}


\title{LPH-VTON: Resolving the Structure-Texture Dilemma of Virtual Try-On via Latent Process Handover} 

\titlerunning{Abbreviated paper title}

\author{Yixin Liu\inst{1}\thanks{Equal contribution.} \and
Baihong Qian\inst{1}\protect\footnotemark[1] \and
Jinglin Jiang\inst{1} \and 
Jeffery Wu\inst{1} \and 
Yan Chen\inst{1} \and 
Wei Wang\inst{1} \and 
Yida Wang\inst{1} \and 
Lanqing Yang\inst{1} \and 
Guangtao Xue\inst{1}}

% --- 机构部分 ---
\institute{Shanghai Jiao Tong University \\
} 
\maketitle
% \begin{figure}

%     \centering
%     \includegraphics[width=\textwidth]{fig/overall.pdf}
%     % \captionof{figure}{Virtual try-on performance of \model{}. By unifying the strengths of both robust structure and vibrant texture within a single generative process, \model{} achieves a new state of the art in both realism and coherence.}
%         \caption{Virtual try-on performance of \model{} on Dresscode\cite{he2024dresscode} and Fashionpedia\cite{jia2020fashionpedia}. Our model demonstrates exceptional robustness and generalization, achieving superior performance in both constrained in-shop environments and unconstrained in-the-wild settings.}
%     \label{fig:overallperform}
% \end{figure}

\begin{figure}[htb]
    \centering
    % width 设为 \textwidth 即可填满单栏宽度
    \includegraphics[width=\textwidth]{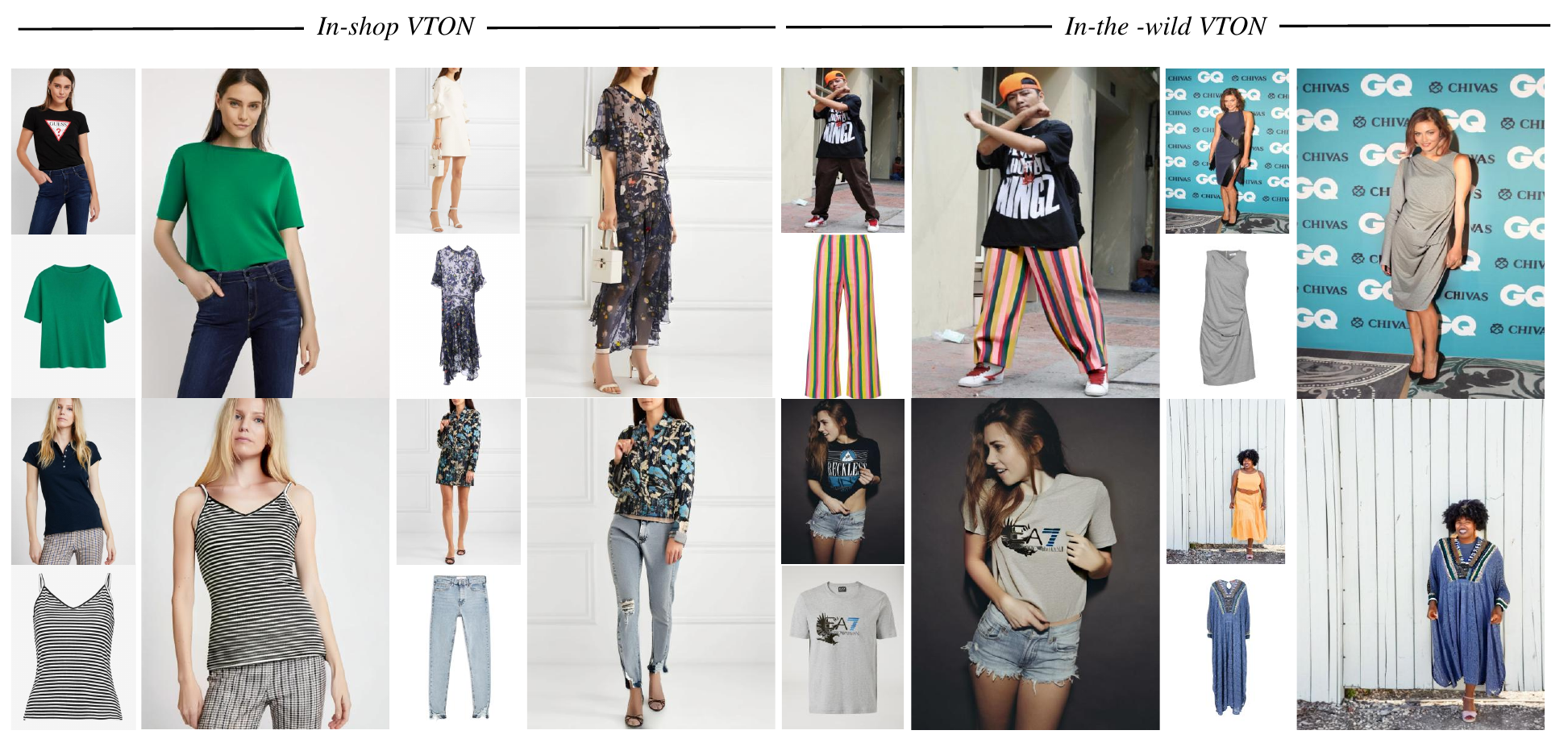}
    \caption{Virtual try-on performance of \model{} on Dresscode\cite{he2024dresscode} and Fashionpedia\cite{jia2020fashionpedia}. Our model demonstrates exceptional robustness and generalization, achieving superior performance in both constrained in-shop environments and unconstrained in-the-wild settings.}
    \label{fig:overallperform}
\end{figure}

\begin{abstract}

% Virtual Try-On (VTON) faces a fundamental trade-off between structural integrity and textural fidelity. 
Virtual Try-On (VTON) aims to synthesize photorealistic images of garments precisely aligned with a person's body and pose. Current diffusion-based methods, however, face a fundamental trade-off between structural integrity and textural fidelity. 
%
% In this paper, we are the first to formalize this challenge as a consequence of complementary inductive biases inherent in prevailing architectures: structure-guided models excel at low-frequency geometry but corrupt high-frequency textures, while texture-enhancing models do the inverse. 
In this paper, we formalize this challenge as a consequence of complementary inductive biases inherent in prevailing architectures: models heavily reliant on spatial constraints naturally favor geometric alignment but often suppress textures, whereas models dominated by unconstrained generative priors excel at vibrant detail rendering but are prone to structural drift. 
Based on this diagnosis, we propose \model{}, a new synergistic framework that resolves this tension within a single, continuous denoising process. \model{} strategically decomposes the generation, leveraging a structure-biased model to establish a geometrically consistent latent scaffold in the early stages, before handing over control to a texture-biased model for high-fidelity detail rendering.
% 这两种模型可能不适合说得这么绝对，我推荐bias，因为实际上都是完整的模型，只是某些训练策略和数据集、模型架构导致的偏好，并不是专门完全为某方面设计的
%
%
% The handover between incompatible backbones  is enabled by a principled interface comprising three core mechanisms: on-the-fly Latent Space Harmonization, a lightweight Latent Adapter, and Trajectory Extension. 
%
Extensive experiments validate our approach. Our model achieves a superior Pareto-optimal balance, establishing new benchmarks in perceptual faithfulness while maintaining highly competitive structural alignment across the standard dataset VITON-HD, proving the efficacy of temporal architectural decoupling.

% Our resulting model, \model{}, establishes a new state-of-the-art, outperforming the previous SOTA methods in SSIM and LPIPS on  both VITON-HD and DressCode datasets. \model{} synthesizes results that are simultaneously structurally coherent and texturally photorealistic, validating our analysis and proving the efficacy of our proposed framework.

\end{abstract}

\section{Introduction}
\label{sec:intro}

Virtual Try-On (VTON) enhances both online and offline shopping experiences by synthesizing authentic garment images on arbitrary human models~\cite{han2018viton, wang2018toward}. Recently, diffusion-based methods have significantly advanced the state of the art in VTON photorealism~\cite{zhu2023tryondiffusion, gou2023dci, morelli2023ladi, kim2024stableviton}. 

Despite this progress, specific architectural choices inevitably imbue models with distinct inductive biases, leading to a fundamental trade-off between structural integrity (spatial alignment) and textural fidelity (photorealistic details). For instance, spatial-concatenation models like CatVTON~\cite{chong2024catvton} exhibit exceptional robustness in aligning complex poses. However, this heavy reliance on spatial constraints favors low-frequency convergence, often yielding overly smooth or rigidly 2D-like textures. 

\begin{figure}[t]
    \centering
    \includegraphics[width=0.7\linewidth]{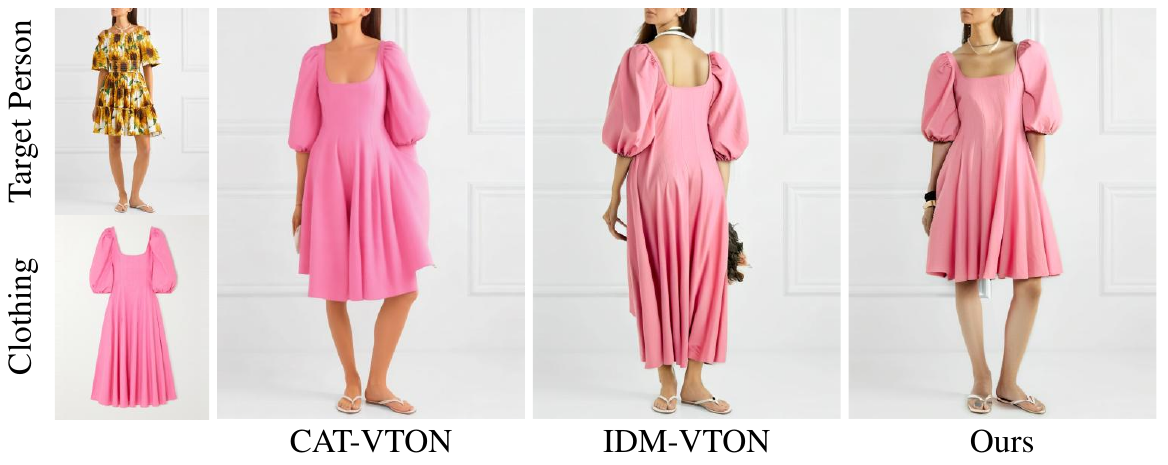} 
    % \caption{An example of the Structure-Texture Trade-off in VTON. Models with a structural bias often lack high-frequency details (e.g., simplified skirt texture), while models with a strong generative prior may fail structurally (e.g., hallucinating folds or inverting foreground/background). Our proposed \model{} resolves this trade-off, achieving Pareto-optimal structural integrity and textural fidelity.}
    \caption{An example of the Structure-Texture Trade-off in VTON. CatVTON yields flat, overly smoothed textures. Conversely, IDM-VTON suffers from severe structural drift, drastically altering the original pose and skirt length. Our \model{} successfully resolves this, generating photorealistic details while strictly preserving accurate spatial geometry.}
    \label{fig:trade-off}
\end{figure}

Conversely, models leveraging cross-attention mechanisms, such as IDM-VTON~\cite{choi2024improving}, unlock powerful generative priors. They excel at producing vibrant, intricate textures but often sacrifice local structural faithfulness. Lacking rigid spatial bounds, they are prone to "semantic drift," hallucinating incorrect fabric folds or distorting the garment's topology. Thus, monolithic architectures inherently favor either geometric stability or textural richness, leaving a gap for a truly comprehensive solution (\cref{fig:trade-off}).

This observation motivates a paradigm shift away from monolithic designs. We formalize this conflict as a direct consequence of complementary inductive biases: structural constraints secure robust geometry but suppress high-frequency textures, while generative priors enhance vibrant details but remain structurally fragile. 

Rather than forcing a single model to mathematically balance these competing biases, we propose Latent Process Handover (LPH), a novel framework that resolves this tension through temporal decoupling. LPH initiates the generation with a structure-biased model to establish a reliable geometric scaffold in the early denoising stages. Once secured, LPH smoothly transitions the latent state to a texture-enhancing model. Tamed by the established geometry, this second model synthesizes high-fidelity details without structural hallucination. This heterogeneous handover is enabled by a parameter-efficient Latent Adapter and a noise-injection step to restore generative plasticity. 

Applying this framework, our resulting model, \model{}, demonstrates the success of this approach by synthesizing results that are simultaneously structurally coherent and texturally photorealistic.

In summary, our work makes three interconnected contributions:
First, we provide a profound architectural diagnosis of the long-standing structure-texture conflict, formalizing it as a consequence of the complementary inductive biases inherent in spatial concatenation and cross-attention priors. 
Second, we propose Latent Process Handover (LPH), a novel generative framework that strategically decomposes the diffusion trajectory. By bridging heterogeneous backbones via a parameter-efficient Latent Adapter and a re-noising process, LPH enables the temporal decoupling of structural anchoring and textural enhancement. 
Finally, extensive experiments demonstrate the superiority of our approach. Rather than over-optimizing for global image distributions at the expense of local details, \model{} achieves a Pareto-optimal balance. It establishes a new benchmark in perceptual faithfulness (LPIPS/SSIM) while maintaining highly competitive generative realism, offering a robust solution for authentic virtual try-on.
\section{Related Work}
\label{sec:related}

\subsection{Generative Virtual Try-On}
Early Virtual Try-On (VTON) systems applied GAN-based methods~\cite{goodfellow2014generative, choi2021viton}, which often struggled with blurring, warping artifacts, and generalization limitations ~\cite{lee2022high, ge2021parser}. For instance, GP-VTON~\cite{xie2023gp} is a notable GAN-based method that achieves competitive performance through collaborative local-flow and global-parsing learning, demonstrating the upper limits of the GAN paradigm. More recently, VTON systems have evolved to diffusion-based architectures ~\cite{song2025survey}, using Latent Diffusion Models (LDMs)~\cite{ho2020denoising, rombach2022high, dhariwal2021diffusion} to enable superior photorealism and stability. Recent diffusion-based models, such as TryOnDiffusion~\cite{zhu2023tryondiffusion}, LaDI-VTON~\cite{morelli2023ladi}, and StableVITON~\cite{kim2024stableviton}, have significantly advanced the state of the art. However, within these powerful generative models, a fundamental tension between structural coherence and textural fidelity has emerged, driven by two underlying architectural philosophies.

\textbf{Attention-Guided Models.} Frameworks like IDM-VTON~\cite{choi2024improving} and StableVITON~\cite{kim2024stableviton} use cross-attention mechanisms to inject garment features into the generative process. This architecture excels at rendering vibrant textures due to the nature of cross-attention, which directly transfers high-frequency appearance details. However, it often does so at the cost of geometric consistency, resulting in structural distortions, semantic drift, and other hallucinated artifacts.

\textbf{Architecturally-Constrained Models.} In contrast, methods like CAT-VTON~\cite{chong2024catvton} and ControlNet~\cite{zhang2023adding} incorporate strong geometric priors through explicit spatial conditioning, akin to early spatial concatenation methods. This design enforces stable and accurate structural alignment; however, the heavy reliance on rigid spatial constraints limits the model’s ability to capture high-frequency appearance cues, frequently leading to overly smooth, rigid, or texture-deficient results. 

Our work aims to unify these complementary approaches by exploring compositional strategies.

\subsection{Compositional Generative Modeling}
Model composition plays a key role in the development of generative models, wherein multiple model components are combined to leverage complementary strengths. Existing general-purpose approaches include pre-inference weight-space operations like model averaging~\cite{wortsman2022model} or merging~\cite{yadav2023ties}, and post-generation sequential pipelining, where a completed image is refined by another model~\cite{ho2022cascaded, podell2023sdxl}.

These methods are ill-suited for the specific structure-texture challenge, as they either statically combine models or are constrained by the flaws of an initial complete render. Our \model{} framework introduces a novel form of intra-process composition. This aligns with recent interest in procedural control~\cite{garipov2023compositional, avrahami2023blended} but is uniquely tailored to harness distinct model strengths for VTON.
\section{Methodology}
\label{sec:method}

% 我们的研究动机源于虚拟试穿（VTON）中一个根本性的权衡：结构一致性与纹理真实感之间的矛盾。我们提出了潜在过程切换（Latent Process Handover, LPH）框架，这是一个全新的范式。它并非通过构建一个庞大的单一模型来解决此问题，而是在一个单一、连续的生成过程中，协同地组合两个专门的、异构的扩散模型。我们通过策略性地划分去噪轨迹，并引入一个有原则的切换机制来桥接这两个模型，从而实现这一目标。我们的\model{}框架概览如图\ref{fig:framework}所示。

\begin{figure}[t]
  \centering
  \includegraphics[width=\linewidth]{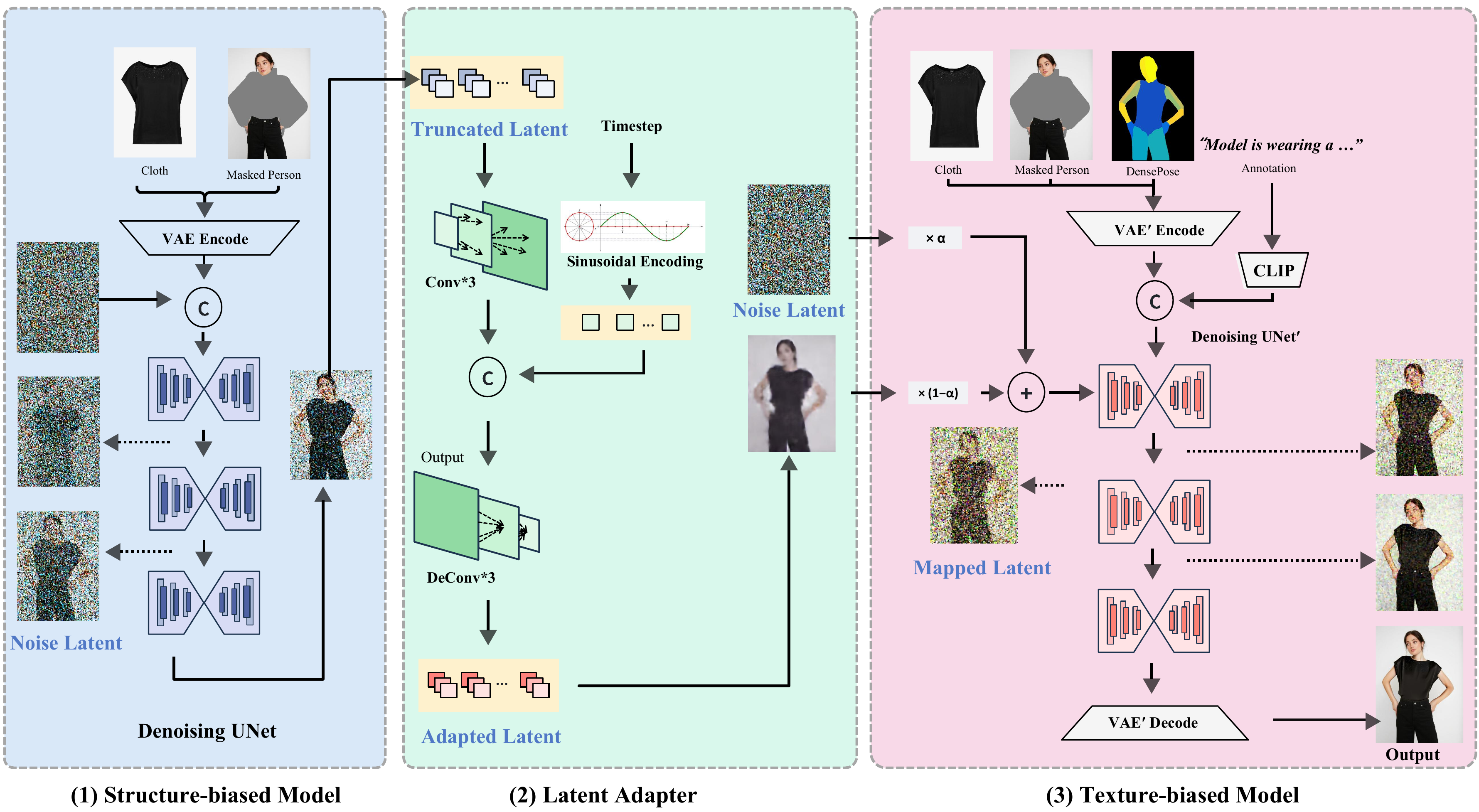} % Your system diagram image
  \caption{\textbf{Overview of our \model{} Framework.} Our method orchestrates a two-phase denoising process. \textbf{Phase 1:} A Structure-biased Model uses minimal inputs (Cloth, Masked Person) to generate a geometrically sound latent scaffold. \textbf{Handover:} At a designated timestep, the core \textbf{Latent Adapter} translates this intermediate state to bridge the distributional gap. \textbf{Phase 2:} A Texture-biased Model, conditioned on richer inputs (e.g., text, DensePose), takes over to render a high-fidelity, photorealistic image.}
  \label{fig:framework}
\end{figure}

Our research is motivated by a fundamental trade-off in virtual try-on (VTON) between structural coherence and textural realism. We propose the LPH framework, a novel paradigm that resolves this dichotomy not by creating a new monolithic architecture, but by synergistically composing two specialized, heterogeneous diffusion models within a single, continuous generative process.
This is achieved by strategically partitioning the denoising trajectory and introducing a principled handover mechanism to bridge the models. An overview of our \model{} architecture is depicted in \cref{fig:framework}.

% 去噪扩散概率模型（DDPMs）的迭代特性为过程控制提供了天然的基础。潜在扩散模型（LDM）\cite{Rombach2022}通过逆转一个前向加噪过程来工作。这个逆向过程是一系列去噪步骤，其中在时间步 t−1的潜变量状态是从时间步 t的状态估计得出的。DDPM采样器的一个通用单步公式为：
% 个逐步迭代的公式揭示了最终输出z0是一个马尔可夫链的结果。这种结构在任何时间步t都提供了天然的干预点（intervention points）。我们不必让单一模型 ϵ θ指导整个轨迹，而是在流程中途改变它，这在理论上是完全合理的。我们可以修改潜变量状态z t ，改变条件c ，或者，正如我们所提议的，将引导模型从 ϵ θ 1    切换到ϵ θ 2。我们的LPH框架正是建立在这一原则之上，利用这些干预点来精心策划一次在具有互补归纳偏置的模型之间的“交棒”。

\subsection{Motivation}
%The Structure-Texture Dilemma in VTON
\label{subsec:motivation}
%结构纹理定义
In the context of virtual try-on, we define \emph{Structure} as the rigorous spatial alignment to the target body mask and pose. Conversely, \emph{Texture} refers to the microscopic photorealistic details, including high-frequency fabric patterns and natural lighting. Current monolithic diffusion models inherently struggle to optimize both simultaneously. We observe that this dilemma is not merely an artifact of insufficient training, but is deeply rooted in the inductive biases of their underlying architectural frameworks.

%拿我们使用的两个模型作为例子
Taking two representative state-of-the-art models as examples: CatVTON, due to its spatial concatenation design, exhibits strong geometric constraints but oversmoothed textures; while IDM-VTON, leveraging cross-attention and the rich prior of SDXL, achieves vivid textures yet suffers from geometric instability. We emphasize that these two models serve as examples; other models may exhibit similar biases through different mechanisms, but the underlying trade-off remains pervasive. A deeper theoretical analysis of bias-variance in VTON models is provided in the supplementary material.

This diagnosis motivates our proposal to harness their complementary strengths via a compositional generation framework, rather than seeking a single architecture that must inevitably compromise.

\subsection{Preliminaries}
%Diffusion Models as Procedural Generators
Recent SOTA virtual try-on methods predominantly build upon diffusion models~\cite{song2025survey}. Denoising Diffusion Probabilistic Models (DDPMs)~\cite{ho2020denoising} generate data by reversing a forward noising process. Latent diffusion models (LDMs)~\cite{rombach2022high} operate in the compressed latent space of a VAE for efficiency. The reverse process is a Markov chain that progressively denoises a random latent \(z_T \sim \mathcal{N}(0,I)\) to a clean latent \(z_0\):

\begin{equation}
z_{t-1} = \frac{1}{\sqrt{\alpha_t}} \left( z_t - \frac{1-\alpha_t}{\sqrt{1-\bar{\alpha}_t}} \epsilon_\theta(z_t, t, c) \right) + \sigma_t \epsilon,
\label{eq:ddpm}
\end{equation}

where \(\epsilon_\theta\) is the noise-prediction network, \(c\) is conditioning (e.g., garment image, mask, pose), and \(\alpha_t, \bar{\alpha}_t, \sigma_t\) define the noise schedule.

Crucially, this step-by-step formulation reveals that the final output $z_0$ is the result of a Markov chain~\cite{douc2018markov}. This structure offers natural intervention points at any timestep $t$. Instead of a single model $\epsilon_\theta$ guiding the entire trajectory, it is theoretically sound to alter the process mid-stream. One can modify the latent state $z_t$, change the conditioning $c$, or, as we propose, \emph{switch the guiding model} from $\epsilon_{\theta_1}$ to $\epsilon_{\theta_2}$. Our \model{} framework is built upon this principle, leveraging these intervention points to orchestrate a handover between models with complementary inductive biases.

\subsection{Framework Overview}
% Latent Process Handover: Overview
\label{subsec:overview}
As illustrated in \cref{fig:framework}, we partition the entire denoising process of $T$ steps into two distinct yet continuous phases, orchestrated by a central handover mechanism.

\textbf{Phase 1: Structure-Guided Scaffolding (Steps $T \to T-T_s$).} The first phase prioritises structural integrity over textural detail. We employ a structure-biased model conditioned on a minimal set of structural cues, $c_S = \{I_{gar}, I_{masked}\}$. As shown in \cref{fig:framework}, the target cloth image ($I_{gar}$) and the masked person image ($I_{masked}$) are encoded by a VAE encoder $\mathcal{E}$ to provide the necessary geometric constraints. Starting from random noise $z_T$, the model $\epsilon_{\theta_S}$ iteratively denoises the latent for $T_s$ steps. The output of this phase is the \emph{Truncated Latent} $z_{T-T_s}$, a state that robustly encodes the overall composition but may lack high-frequency texture information.
% We initialize the process with a \textit{Structure-biased Model}, $\epsilon_{\theta_S}$. For the initial $T_s$ steps, this model focuses exclusively on establishing a robust geometric foundation, ensuring correct garment shape, drape, and pose alignment.

\textbf{Latent Process Handover (Step $T-T_s$).} 
Directly transferring control from \(\epsilon_{\theta_S}\) to \(\epsilon_{\theta_T}\) at step \(T-T_s\) is non-trivial due to their incompatible latent spaces and the low-entropy state of the structure model's latent\cite{yang2023diffusion}. To address this, we introduce two key components: a Latent Adapter that aligns the distributions via learned translation, and a Trajectory Extension mechanism that reintroduces controlled noise to restore generative capacity. These components, detailed in \cref{subsec: adapter}, enable a smooth and effective handover, allowing the texture model to leverage its full potential without compromising the established structural scaffold.
% The Latent Process Handover between the two models with different latent manifolds enables a smooth shift from structure-preserving generation to texture-focused refinement. 

\textbf{Phase 2: Texture-Enhancing Refinement (Steps $T-T_s \to 0$).} 
% Following the handover at step $T-T_s$, the process is seamlessly transferred to a powerful \textit{Texture-biased Model}, $\epsilon_{\theta_T}$. For the remaining $T_t = T-T_s$ steps, this model leverages its superior synthesis capabilities to render fine-grained details, vibrant colors, and realistic fabric textures.
With the latent state successfully handed over and prepared, the process control is transferred to the texture-biased model $\epsilon_{\theta_T}$. This model is conditioned on a richer set of inputs $c_T = \{I_{gar}, I_{masked}, I_{densepose}, P_{text}\}$, including textual annotations and fine-grained DensePose maps. These additional conditions enable $\epsilon_{\theta_T}$ to leverage the powerful priors of its large-scale backbone for superior textural rendering. Starting from the prepared latent, the model denoises for $T_t$ steps, where $T_t \geq T-T_s$ denotes the number of Phase~2 denoising steps (see \cref{subsec:adapter}). Once the process completes at $t=0$, the final clean latent $z_0$ is decoded by the corresponding VAE decoder $\mathcal{D}'$ into the output image $I_{out} = \mathcal{D}'(z_0)$.

% \subsection{Phase 1: Structure-Guided Latent Scaffolding}
% The first phase prioritizes structural integrity over textural detail. We employ a structure-biased model conditioned on a minimal set of structural cues, $c_S = \{I_{gar}, I_{masked}\}$. As shown in \cref{fig:framework}, the target cloth image ($I_{gar}$) and the masked person image ($I_{masked}$) are encoded by a VAE encoder $\mathcal{E}$ to provide the necessary geometric constraints. Starting from random noise $z_T$, the model $\epsilon_{\theta_S}$ iteratively denoises the latent for $T_s$ steps. The output of this phase is the Truncated Latent $z_{T-T_s}$, a state that robustly encodes the overall composition but may lack high-frequency texture information.

\subsection{Latent Process Handover Mechanism}
\label{subsec:adapter}
% The handover is the technical core of our framework, enabling a smooth transition between two incompatible backbones (e.g., SD-1.5~\cite{Rombach2022}  and SDXL~\cite{Podell2023}) that operate on different latent manifolds, $\mathcal{Z}_S$ and $\mathcal{Z}_T$.

\subsubsection{Latent Adapter for Distributional Alignment.}
To bridge the distributional gap between $\mathcal{Z}_S$ and $\mathcal{Z}_T$, we introduce a lightweight Latent Adapter, $\mathcal{A}_{\phi}$. This module functions as a learned translator that maps the latent representation from the source model's manifold to the target's. As seen at the bottom of \cref{fig:framework}, the adapter takes the Truncated Latent $z_{T-T_s}$ and the current timestep $T-T_s$ as input. The timestep is first converted into a high-dimensional embedding via sinusoidal positional encoding, $\mathrm{pe}(T-T_s)$, to make the adapter aware of the current noise level. The adapter, implemented as a compact U-Net with three down-sampling and up-sampling convolutional blocks, then performs the transformation:
\begin{equation}
    \hat{z}_{T-T_s} = \mathcal{A}_{\phi}(z_{T-T_s}, \mathrm{pe}(T-T_s))
\end{equation}
The resulting \textit{Adapted Latent} $\hat{z}_{T-T_s}$ is now statistically aligned with the distribution that the texture-biased model expects at that specific stage of denoising.

\subsubsection{Trajectory Extension for Generative Potential.}

We empirically observed that directly handing over the adapted latent \( \mathcal{A}_\phi(z_{T-T_s}^{(S)}) \) to the texture-biased model often yields muted colors and flattened textures. This occurs because the structure model's latents at handover have low conditional entropy, 
 severely limiting the texture model's generative capacity. To restore generative freedom while preserving the established structure, we propose Trajectory Extension—a simple but effective technique that increases the number of denoising steps allocated to the second phase.

% \begin{equation}
% H(z_{t-1}|z_t) = -\int p(z_{t-1}|z_t)\log p(z_{t-1}|z_t) dz_{t-1},
% \label{eq:entropy}
% \end{equation}

% which drops to 4.23 without intervention,

Concretely, given a target total of \(T=30\) steps and a handover point \(T_s\), we let the structure model run for \(T_s\) steps, producing \(z_{T-T_s}^{(S)}\). We denote a configuration as \((T_s,\,T_t)\), where $T_t \geq T-T_s$ is the number of denoising steps allocated to Phase~2. Instead of starting the texture model from the handover timestep $T-T_s$, we ``rewind'' the process by initializing it at an earlier, higher-noise timestep $T_t$. For instance, extending a \((18,12)\) configuration to \((18,18)\) sets the texture model's starting timestep to \(18\) and allows it to denoise for 18 steps. In practice, this is efficiently implemented by feeding the adapted latent \(\mathcal{A}_\phi(z_{T-T_s}^{(S)})\) into the texture model's sampler with a corresponding denoising strength parameter, which instructs the sampler to first add noise up to timestep $T_t$ before proceeding with the reverse process.
%

% \subsection{Phase 2: Texture-Enhancing Refinement}
% With the latent state successfully handed over and prepared, the process control is transferred to the texture-biased model $\epsilon_{\theta_T}$. This model is conditioned on a richer set of inputs $c_T = \{I_{gar}, I_{masked}, I_{densepose}, P_{text}\}$, including textual annotations and fine-grained DensePose maps. These additional conditions enable $\epsilon_{\theta_T}$ to leverage the powerful priors of its large-scale backbone for superior textural rendering.

% Starting from the prepared latent, the model denoises for the remaining $T_t$ steps. Once the process completes at $t=0$, the final clean latent $z_0$ is decoded by the corresponding VAE decoder $\mathcal{D}'$ into the final, high-fidelity \textit{Output Image} $I_{out} = \mathcal{D}'(z_0)$.

\subsection{Training Strategy}
A key advantage of our LPH framework is its training efficiency. Both large backbone models, $\epsilon_{\theta_S}$ and $\epsilon_{\theta_T}$, remain frozen. We only train the lightweight Latent Adapter $\mathcal{A}_{\phi}$. We first curate a dataset of paired latent vectors by running both models on the same image data (using their respective conditioning sets $c_S$ and $c_T$) for a full denoising trajectory. This yields pairs of latent states $(z_t^{(S)}, z_t^{(T)})$ at every timestep $t$. The adapter is then trained via a direct regression objective to minimize the Mean Squared Error (MSE) loss:
\begin{equation}
    \mathcal{L}_{\text{Adapter}} = \mathbb{E}_{z_t^{(S)}, z_t^{(T)}, t} \left[ \left\| \mathcal{A}_{\phi}(z_t^{(S)}, \mathrm{pe}(t)) - z_t^{(T)} \right\|_2^2 \right]
\end{equation}
This objective efficiently teaches the adapter to perform the precise mapping required to bridge the distributional gap between the two models across all stages of the generative process.
\section{Experiments}
\label{sec:experiments}

%-------------------------------------------------------------------------
\subsection{Experimental Setup}

\textbf{Datasets and Backbones.}
Our experiments are conducted on two standard high-resolution benchmarks: VITON-HD~\cite{choi2021viton} and DressCode~\cite{he2024dresscode}, evaluated at a 1024x768 resolution. Our LPH framework is instantiated with two powerful, publicly available backbones. For the Structure-Guided model, we employ CatVTON~\cite{chong2024catvton}, built on Stable Diffusion 1.5. For the Texture-Enhancing model, we use IDM-VTON~\cite{choi2024improving}, which leverages the SDXL backbone.

\noindent \textbf{Baselines.}
We provide a comprehensive comparison against state-of-the-art methods that represent diverse architectural paradigms, including OOTDiffusion~\cite{xu2025ootdiffusion}, StableVITON~\cite{kim2024stableviton}, and DCI-VTON~\cite{gou2023dci}. To validate our claims about complementary biases, we also explicitly evaluate the standalone performance of our backbone models, CatVTON and IDM-VTON, which serve as critical points of reference in our ablation studies.

\noindent \textbf{Evaluation Metrics.}
We employ a suite of widely-used metrics to assess both structural accuracy and perceptual realism. For image-level fidelity on paired data, we report the Structural Similarity Index (SSIM)$\uparrow$ and Learned Perceptual Image Patch Similarity (LPIPS)$\downarrow$. For evaluating the realism and distribution similarity on unpaired data, we use the Fréchet Inception Distance (FID)$\downarrow$ and Kernel Inception Distance (KID)$\downarrow$. Higher SSIM is better, while lower values are better for all other metrics.

\noindent \textbf{Training Details.}
Our training is highly efficient as it does not require training large models from scratch. The pre-trained  backbones are kept frozen. We first train only the lightweight Latent Adapter ($\mathcal{A}_\phi$) on a pre-computed dataset of paired latent vectors, as detailed in \cref{subsec:adapter}. This phase converges quickly. For optimal results, we subsequently perform a brief end-to-end fine-tuning of the entire pipeline with a small learning rate. The entire training process was conducted on 2 NVIDIA L20 GPUs with a batch size of 8.

\begin{table}[t]
\centering
\caption{Quantitative comparison with state-of-the-art methods on VITON-HD. The best-performing method is \textbf{bolded}, and the second-best is \underline{underlined}.}
\label{tab:main_results}

\setlength{\tabcolsep}{6pt} % 列数减少了，可以稍微放大列间距
\small
\renewcommand{\arraystretch}{1.2}

\begin{tabular}{l cccc cc}
\toprule
\multirow{3}{*}{\textbf{Method}} & \multicolumn{4}{c}{Paired} & \multicolumn{2}{c}{Unpaired} \\
\cmidrule(lr){2-5} \cmidrule(lr){6-7}
& SSIM$\uparrow$ & LPIPS$\downarrow$ & FID$_p$$\downarrow$ & KID$_p$$\downarrow$ & FID$_u$$\downarrow$ & KID$_u$$\downarrow$ \\
% 单位行：利用 tiny 字体标注缩放因子
& & & & \scriptsize{($\times 10^{-3}$)} & & \scriptsize{($\times 10^{-3}$)} \\
\midrule

GP-VTON          & \textbf{0.8883} & \underline{0.0824} & 12.27 & 3.94 & \textbf{8.34} & 4.42 \\
DCI-VTON         & 0.8368 & 0.0944 & \underline{9.13}  & \textbf{0.69} & 8.62 & 3.54 \\
Ladi-VTON        & 0.8425 & 0.1451 & 14.77 & 3.56 & 12.55 & 5.61 \\
Stable-VTON-1024 & 0.8646 & 0.1198 & 13.78 & 4.48 & 10.67 & 6.42 \\
% Leffa            & 0.8564 & 0.1052 & \textbf{5.90}  & \underline{0.85} & \underline{8.55} & \textbf{0.96} \\
CatVTON    & 0.8041 & 0.1437 & 14.29 & 2.79 & 9.78  & 2.87 \\
IDM-VTON   & 0.8502 & 0.0875 & 16.13 & 2.50 & 12.13 & 2.66 \\

\midrule
\textbf{Ours}    & \underline{0.8719} & \textbf{0.0791} & 12.05 & 2.21 & 9.71 & \underline{2.56} \\

\bottomrule
\end{tabular}
\end{table}

\begin{table}[t]
\centering
%\vspace{-0.1in}
\caption{Computational Cost Analysis. All metrics are measured for a single image generation at 1024x768 resolution on an NVIDIA A100 40G GPU.}
\label{tab:cost_analysis}
\resizebox{0.8\linewidth}{!}{
\begin{tabular}{lccc}
\toprule
\textbf{Method} & \textbf{TFLOPS} & \textbf{Inference Time (s)} & \textbf{Peak GPU Memory (MB)} \\
\midrule    
CatVTON  & 116.92 & 8.61 & 6606 \\
IDM-VTON  & 156.48 & 8.53 & 16306 \\
\textbf{Ours} & \textbf{140.37} & \textbf{8.70} & \textbf{22312} \\
\bottomrule
\end{tabular}
}
%\vspace{-0.2in}
\end{table}

\subsection{Quantitative Comparison.}

\subsubsection{Effect Comparison.}
\cref{tab:main_results} presents the quantitative results on the VITON-HD dataset. Our \model{} demonstrates a superior Pareto-optimal balance between structural integrity and perceptual realism. Notably, we consistently outperform our direct monolithic diffusion baselines (IDM-VTON and CatVTON) across all evaluated metrics, validating the efficacy of our temporal decoupling strategy. When compared to specialized architectures, the results explicitly reflect the perception-distortion tradeoff. For instance, the warping-based GP-VTON achieves high SSIM through strict pixel preservation but fundamentally bottlenecks natural texture synthesis, lagging significantly in LPIPS and generative metrics (FID). Conversely, the prior-heavy DCI-VTON attains strong global distribution matching (FID) at the severe cost of fine-grained local fidelity (inferior LPIPS). 

Our framework effectively bridges this gap, achieving the best LPIPS score while maintaining highly competitive SSIM and FID. Comprehensive qualitative comparisons are provided in the Supplementary Material.

\subsubsection{Efficiency Comparision.}
As presented in \cref{tab:cost_analysis}, our Latent Process Handover method introduces a marginal increase in inference time compared to the baselines, demonstrating that the handover mechanism itself is highly efficient. The primary trade-off of our composite framework is an increase in peak GPU memory usage (22312MB), as the system must hold components from both heterogeneous backbones in memory.
Notably, the total computational cost of our method (140.37 TFLOPS) is positioned between that of CatVTON (116.92 TFLOPS) and a full IDM-VTON inference (156.48 TFLOPS). This reflects the efficiency of our two-phase design, which leverages the larger SDXL-based model for only the final $k_{struct}$ refinement steps, thus avoiding the cost of a full inference pass with the more computationally intensive model. This analysis confirms that our approach achieves a substantial improvement in synthesis quality while maintaining a practical and efficient computational profile.

\begin{figure}[t]
    \centering
    \includegraphics[width=\linewidth]{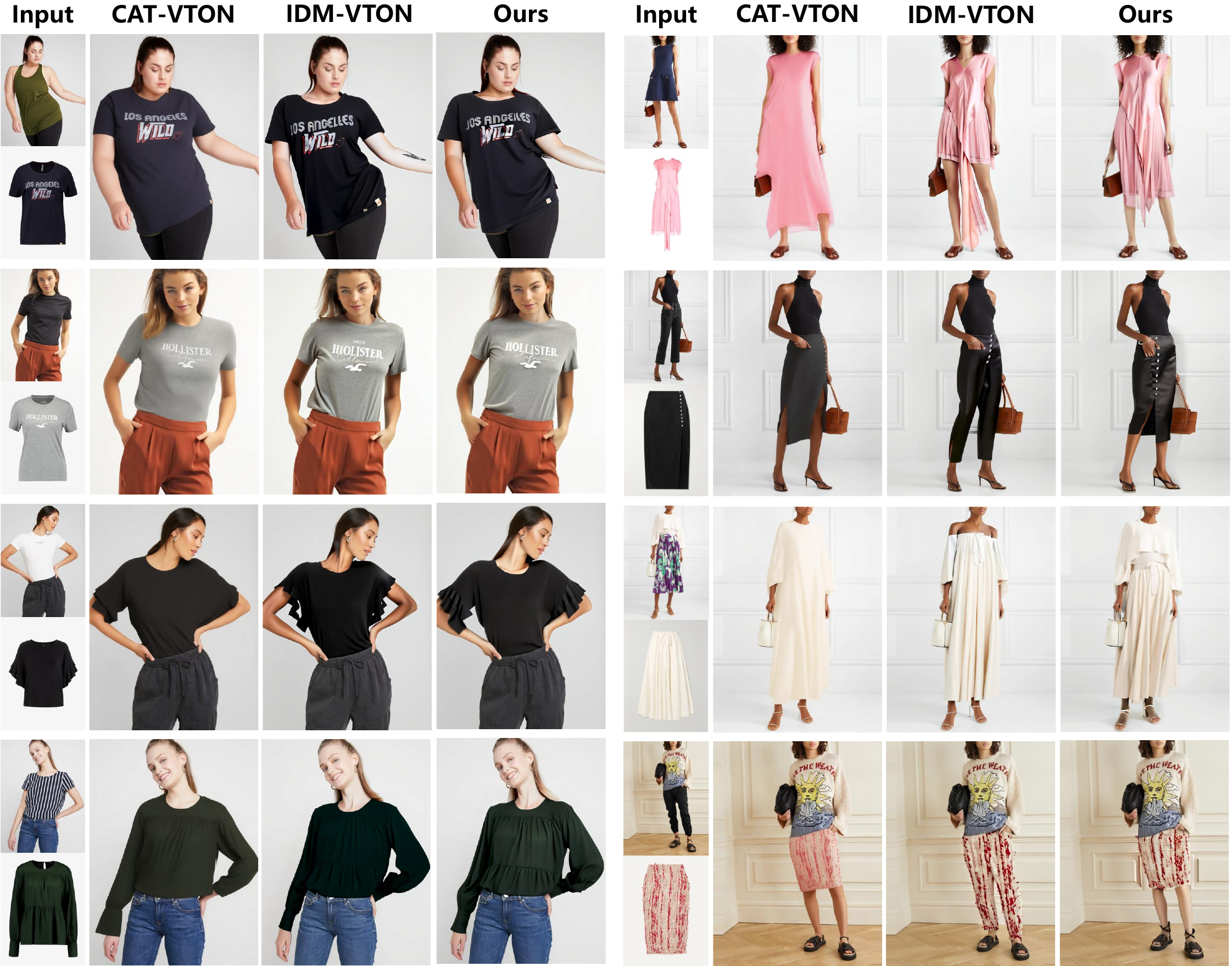} 
    \caption{Qualitative comparison on the DressCode dataset. CatVTON and IDM-VTON demonstrate distinctly different generation biases, while our framework outperforms them both in terms of texture and structure.}
    \label{fig:partofcmp}
\end{figure}
\begin{figure}[t]
    \centering
    \includegraphics[width=\linewidth]{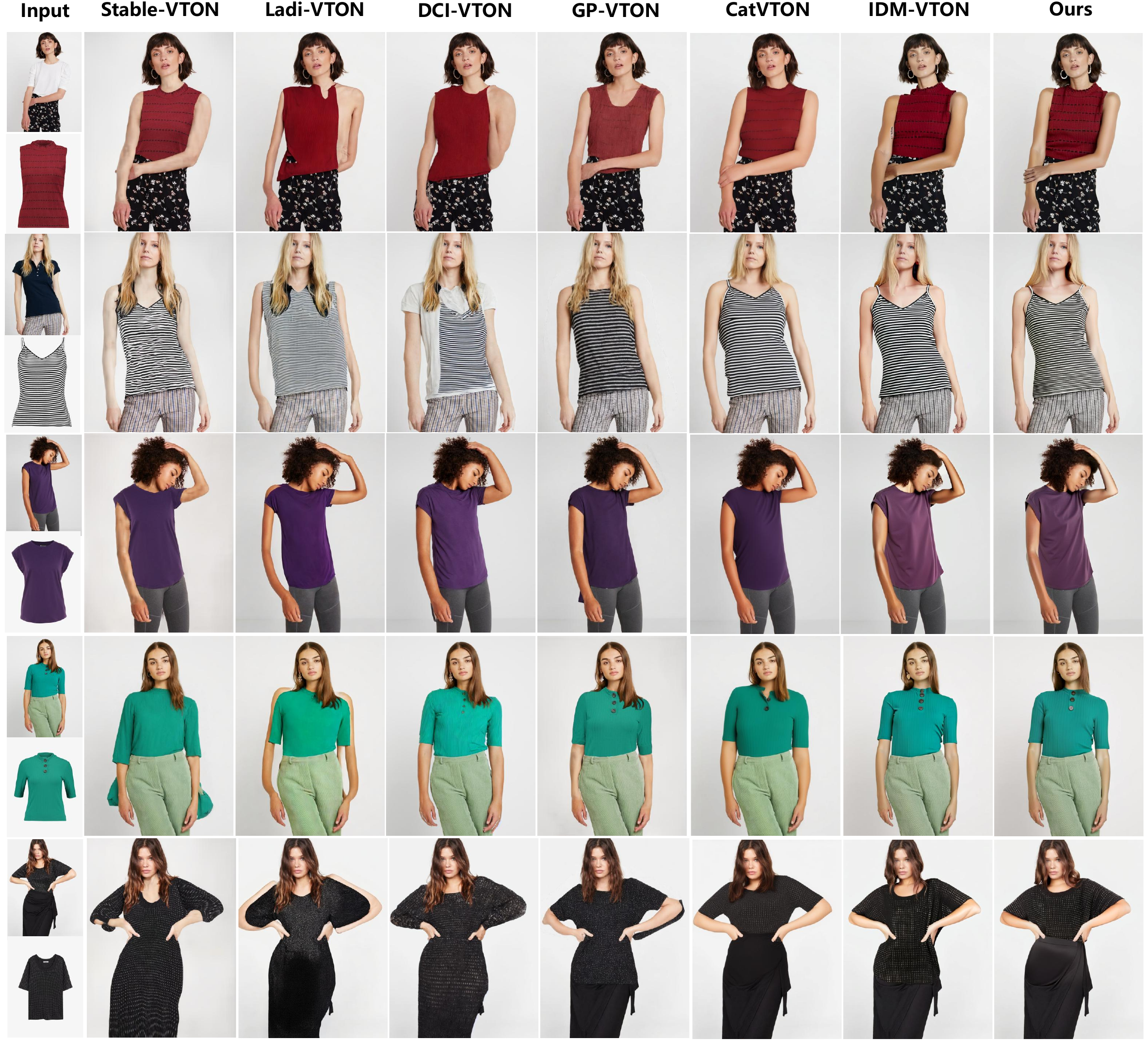} 
    \caption{\textbf{Qualitative Comparison with State-of-the-Art Methods.} We compare LPH-VTON against representative baselines, including StableVITON, LaDI-VTON, DCI-VTON, GP-VTON, CatVTON, and IDM-VTON. While competitors often struggle with either texture blurring or structural artifacts, our method consistently achieves high-fidelity generation with accurate garment details and natural draping.}
    \label{fig:all}
\end{figure}
\subsection{Qualitative Comparison.}

\cref{fig:trade-off} visually corroborates our diagnosis of the generating preferences. Our superior performance is demonstrated in \cref{fig:partofcmp} and \cref{fig:all}. When faced with challenging cases in both settings, the baselines exhibit complementary failure modes rooted in their inductive biases. IDM-VTON, biased towards texture, often fails to preserve the garment's global structure, incorrectly rendering a dress as trousers (e.g., \cref{fig:partofcmp}, column 7). Conversely, CatVTON, biased towards structure, successfully preserves the garment shape but at the cost of severe texture degradation, producing blurry details and unrealistic fabric rendering (e.g., \cref{fig:partofcmp}, column 2 \& 6). In contrast, \model{} successfully synthesizes both the correct garment shape and its high-fidelity texture across all tested in-shop and in-the-wild scenarios, demonstrating the effectiveness of our synergistic LPH framework at resolving this trade-off. In \cref{fig:all}, further comparisons with a broader range of models have validated the effectiveness of our model. For example, in row 4, only our model and GP-VTON correctly produce  the number of buttons. However, our performance in generating striped patterns is clearly superior to GP-VTON(row 2). 

% 

%-------------------------------------------------------------------------
\subsection{Ablation Studies and Analysis}

% In your main .tex file, inside the Ablation Studies subsection

\subsubsection{Analysis of the Handover Point.}
A critical design choice in our \model{} framework is the handover point, which determines the division of labor between the structurally-biased and texture-biased models within a fixed budget of 30 total denoising steps. We parameterize this by $k_{\text{struct}}$, the number of initial steps allocated to the first model, with the remaining $k_{\text{texture}} = max(18, 30 - k_{\text{struct}})$ steps completed by the second model. This parameter governs the fundamental trade-off between the structural integrity of the initial scaffold and the generative freedom afforded to the final refinement stage. To investigate this relationship, we evaluate \model{}'s performance across a range of handover configurations: $(k_{struct}, k_{texture})$ pairs from $(6, 24)$ to $(24, 18)$. 

The results, presented quantitatively in \cref{tab:handoverpoint}, reveal a striking non-monotonic relationship between the handover point and generation quality. The $(0, 30)$ configuration, relying solely on the texture-biased model, often fails to produce coherent structures, resulting in poor performance. As we introduce initial structural steps (e.g., at the $(6, 24)$ configuration), both SSIM and LPIPS metrics improve, demonstrating that leveraging an initial structural bias is vital for forming a reliable geometric anchor.

% Intriguingly, this trend does not continue linearly. At the $(12, 18)$ handover point, we observe a significant and unexpected degradation in performance, with LPIPS and SSIM scores that are substantially worse than those at the $(6, 24)$ configuration. The performance then recovers as $k_{\text{struct}}$ increases further, stabilizing towards the $(30, 0)$ configuration, which relies solely on the structurally-biased model.

% This non-linear "dip" in performance suggests a complex interplay between the two models' generative processes. We hypothesize that at certain intermediate handover points like $(12, 18)$, the latent state exists in a challenging "transitional zone." It may contain just enough structural commitment from the first model to overly constrain the second, while simultaneously lacking the definitive clarity of a more fully-formed scaffold. This can lead to conflicts and inconsistencies when the texture-biased model takes over. This discovery underscores that the handover is not a simple weighted average of model capabilities but a complex orchestration challenge. Our experiments identify an optimal handover window around the $(18, 18)$ configuration, which strikes the most effective balance between establishing a robust structure and enabling vibrant, detailed textural refinement.
Intriguingly, varying the handover timestep $k_{\text{struct}}$ reveals diverging trends across different evaluation metrics rather than a simple linear trajectory. This fluctuation highlights a complex interplay between the contrasting inductive biases of the two architectures. At early handover configurations, the latent state lacks definitive structural clarity, causing the subsequent texture-enhancing model to suffer from semantic drift and geometric collapse. Conversely, as $k_{\text{struct}}$ increases excessively, the structure becomes over-committed; this rigid geometric anchoring stifles the cross-attention mechanisms of the second stage, incrementally degrading textural realism and generative flexibility. Our quantitative analysis identifies the configurations between $(12, 18)$ and $(18, 18)$ as a "Pareto-optimal plateau." Within this specific mid-stage window, the framework achieves an ideal equilibrium: it successfully secures a resilient geometric scaffold while retaining sufficient thermodynamic plasticity. This dynamic orchestration allows the texture-biased network to render high-fidelity, photorealistic details without overriding the established spatial alignment.

\begin{table}[t]
\centering
\caption{Quantitative ablation on the Latent Process Handover configurations. The best-performing method is \textbf{bolded}, and the second-best is \underline{underlined}.}
\label{tab:handoverpoint}

\setlength{\tabcolsep}{3.5pt} % 稍微放宽一点点，3.5pt 在 10 列下通常也能塞进单栏
\footnotesize 
\renewcommand{\arraystretch}{1.1}
\resizebox{\textwidth}{!}{
\begin{tabular}{l l cccc cccc}
\toprule
\multirow{3}{*}{\textbf{Method}} & 
\multirow{3}{*}{\textbf{Steps}} & 
\multicolumn{4}{c}{\textbf{Viton-HD}} & 
\multicolumn{4}{c}{\textbf{DressCode}} \\
% 使用 (lr) 截断横线
\cmidrule(lr){3-6} \cmidrule(lr){7-10}

& & SSIM$\uparrow$ & LPIPS$\downarrow$ & FID$_p$$\downarrow$ & KID$_p$$\downarrow$ & SSIM$\uparrow$ & LPIPS$\downarrow$ & FID$_p$$\downarrow$ & KID$_p$$\downarrow$ \\
% 专门的单位/缩放行
& & & & & {\tiny ($\times 10^{-3}$)} & & & & {\tiny ($\times 10^{-3}$)} \\
\midrule

IDM-VTON & (0,30) & 0.8502 & 0.0875 & 16.12 & 2.50 & 0.9103 & 0.0688 & 15.90 & \underline{3.09} \\
CatVTON  & (30,0) & 0.8041 & 0.1437 & 14.28 & 2.79 & 0.9031 & 0.0795 & 16.48 & \textbf{2.93} \\

\midrule
\multirow{4}{*}{\textbf{Ours}} 
& (6,24)  & 0.8711 & 0.0828 & \textbf{9.31} & 2.56 & 0.9076 & 0.0744 & 10.45 & 4.45 \\
& (12,18) & \textbf{0.8723} & \underline{0.0817} & \underline{9.76} & \underline{2.44} & \underline{0.9125} & \underline{0.0661} & \textbf{9.19} & 4.53 \\
& (18,18) & \underline{0.8719} & \textbf{0.0791} & 12.05 & \textbf{2.21} & 0.9122 & \textbf{0.0641} & \underline{10.25} & 4.01 \\
& (24,18) & 0.8536 & 0.0989 & 11.66 & 2.70 & \textbf{0.9139} & 0.0671 & 11.02 & 4.25 \\

\midrule
RGB Refine. & --- & 0.8397 & 0.0892 & 19.11 & 3.33 & --- & --- & --- & --- \\
\bottomrule
\end{tabular}
}
\end{table}

\begin{figure}[t]
    \centering
    \includegraphics[width=\linewidth]{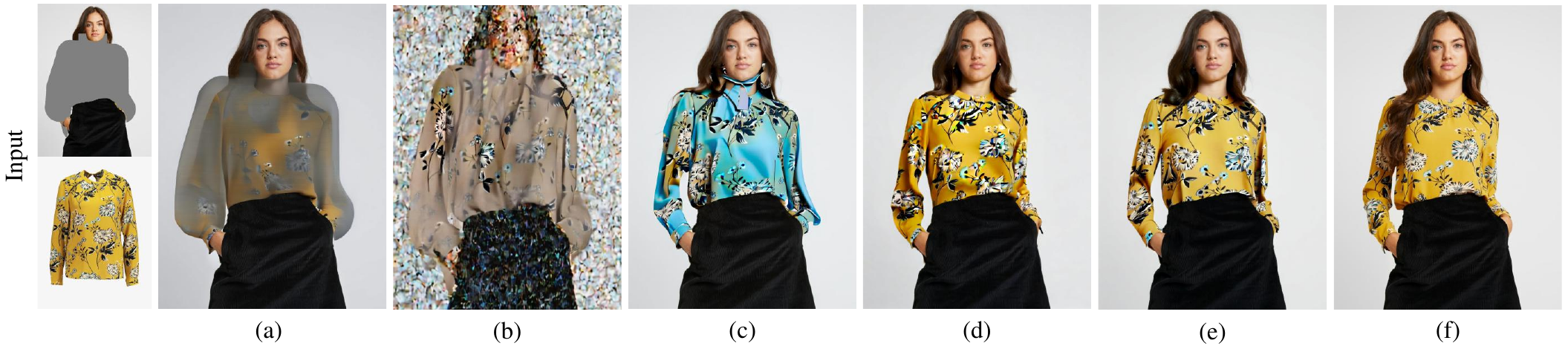} % Replace with your qualitative comparison figure

    \caption{Results of ablation experiments.
    % \textbf{Qualitative comparison of LPH withou a Two-Stage RGB Refinement baseline.}
    (\textbf{a}) The result of a direct handover without Trajectory Extension.
    (\textbf{b}) Handover using RGB pixels instead of latent.
    (\textbf{c}) Latent space handover without the proposed Latent Adapter.
    (\textbf{d}) Two-Stage RGB Refinement. The basic structure looks good, but the patterned color blocks are too large.
    (\textbf{e}) Our generated image. The level of detail and realism in the patterns has been significantly enhanced compared to RGB Refinement.
    (\textbf{f})Real-world wearing photo.
    }
    \label{fig:ablation}
\end{figure}

\subsubsection{Necessity of Trajectory Extension.}
As shown in \cref{fig:ablation}(a), the results of a direct handover suffer from visibly muted colors and a lack of fine-grained textural detail, closely mirroring the textural characteristics of the structurally-biased model. The generated garments appear flatter and less realistic. In contrast, our full \model{} framework with Trajectory Extension(\cref{fig:ablation}(e)) successfully unlocks the synthesis potential of the second model, producing garments with rich color vibrancy and intricate high-frequency textures that align closely with the source garment. This comparison provides clear empirical evidence that the Trajectory Extension is not merely an engineering choice, but a critical component for enabling effective textural refinement in our handover process.

As shown in \cref{fig:ablation}(a), the result of a direct handover without trajectory extension with handover configuration (18,12), the grayscale image of the residual mask is still clearly visible in the final result image, making the clothing section appear gray and hazy. This phenomenon stems from the lack of generating steps of IDM-VTON. Extending the denoising trajectory is critical to our system. Therefore, we extend the IDM-VTON generating steps under 18 to 18.

% \begin{figure}[tbh]
% \centering
% \includegraphics[width=\linewidth]{fig/compare_tex_struc.png}
% \caption{
% \textbf{Ablation study on Trajectory Extension.}
% This figure validates the necessity of our controlled re-noising step.
% (\textbf{a}) The source inputs, including the person and the target garment.
% (\textbf{b}) The result of a direct handover without Trajectory Extension. While structurally coherent, the garment's texture appears muted and lacks vibrancy, inheriting the textural limitations of the first-phase model.
% (\textbf{c}) Our full LPH framework with Trajectory Extension. The re-noising step provides the second-phase model with a higher-entropy starting point, enabling it to synthesize rich colors and fine-grained textural details, which is crucial for photorealism.
% }
% \label{fig:ablation_trajectory}
% \end{figure}

% \begin{figure}[tbp]
%     \centering
%     \includegraphics[width=\linewidth]{fig/compare_tex_struc.png} % Replace with your qualitative comparison figure
%     \caption{\ylq{The Structure-Texture Trade-off in VTON. Texture-biased models like IDM-VTON suffer from structural collapse (e.g., rendering a dress as a shirt), while structure-biased models like CAT-VTON degrade textures and introduce artifacts. Our method resolves this trade-off, achieving both structural integrity and textural fidelity.}}
%     \label{fig:no_adapter}
% \end{figure}

\subsubsection{Necessity of the Latent Adapter.}
Next, we remove the Latent Adapter ($\mathcal{A}_{\phi}$) while retaining the re-noising step. As shown in \cref{fig:ablation}(c), while the model avoids complete geometric collapse, the output is marred by severe visual artifacts and color shifts. This indicates a drastic distributional mismatch between the heterogeneous latent spaces of the structure-biased and texture-biased backbones. This result validates the adapter's critical role as a feature translator, performing the precise alignment necessary to bridge the generative manifolds and maintain a mathematically continuous trajectory during the handover.

% \begin{figure}[h!]
%     \centering
%     \includegraphics[width=\linewidth]{fig/compare_tex_struc.png} % Replace with your qualitative comparison figure
%     \caption{
%     \textbf{Failure case of an in-process pixel-space handover.}
%     This figure illustrates the result of handing over at an intermediate timestep (e.g., t=500) via a latent -> pixel -> latent conversion.
%     (\textbf{a}) The intermediate, noisy RGB image decoded from the first model's latent. It is blurry and lacks detail.
%     (\textbf{b}) The final result after the second model attempts to refine this image. The severe information loss during the conversion leads to incoherent structures and pervasive artifacts, demonstrating the infeasibility of this approach.
%     }
%     \label{fig:rgb_mid}
% \end{figure}

\subsubsection{Necessity of the Latent-Space Handover.}
Our framework's design is centered on a handover within the latent space. To understand why this is critical, we first analyze the fundamental flaws of a handover in pixel space during the generative process. Such a procedure would involve partially denoising with the first model, decoding to an intermediate, noisy RGB image, and then re-encoding this blurry, incomplete image to initialize the second model. This latent -> pixel -> latent conversion acts as a severe information bottleneck, purging high-frequency details and disrupting the continuous generative trajectory essential for coherent synthesis. As qualitatively demonstrated in \cref{fig:ablation}(b), this approach leads to catastrophic failures, yielding incoherent and artifact-ridden results.

We also evaluate a sequential "Two-Stage RGB Refinement" baseline, cascading the fully converged RGB output of the structure-biased model into the texture-biased pipeline. However, this approach also fails due to severe error accumulation. 
Fully denoising the first stage means that any inherent errors, such as structural misalignments or rigid fabric textures, will be passed on to the subsequent stage. Our experimental results demonstrate catastrophic shape deformations and colour shifts (see \cref{fig:ablation}(d)).
\section{Conclusion and Limitations}
\label{sec:conclusion}

In this work, we proposed the Latent Process Handover (LPH) framework, a novel Virtual Try-On (VTON) method that achieves high-fidelity results by explicitly resolving the long-standing trade-off between structural integrity and textural fidelity.
Our method is founded on a novel formulation of the VTON generative process, derived from our systematic analysis attributing this conflict to the complementary inductive biases of existing architectural families. 
To bridge the gap between these competing specialist models within a single, continuous generation, we introduce several key innovations, including a strategic decomposition of the denoising process, a parameter-efficient handover interface, and a lightweight Latent Adapter to align their incompatible latent spaces.
%
% Extensive experiments demonstrate the effectiveness of our approach, significantly outperforming prior state-of-the-art techniques in both structural and perceptual metrics. 
By achieving a superior Pareto-optimal balance, it outperforms prior strong baselines in perceptual faithfulness while maintaining robust geometric alignment. 

While \model{} achieves excellent performance, we acknowledge several limitations that present avenues for future research. The primary limitation is the inference latency of the two-stage pipeline, paving the way for exploration into model distillation to unify the capabilities of both backbones into a single, efficient network. Furthermore, the handover process is currently static, motivating the development of a more sophisticated dynamic routing mechanism that could adaptively determine the optimal handover point based on input complexity. 

% % \section*{Acknowledgements}
% % Please insert your acknowledgments here.

% % ---- Bibliography ----
% %
% % BibTeX users should specify bibliography style 'splncs04'.
% % References will then be sorted and formatted in the correct style.
% %
\bibliographystyle{splncs04}
\bibliography{main}
% \clearpage
\setcounter{page}{1}
% \maketitlesupplementary

\appendix

This supplementary document provides a comprehensive analysis extending the main paper's findings. We first elucidate the specific architectural design and training configurations of the Latent Adapter in \cref{sec:impl_details}. Subsequently, we present an expanded scope of qualitative comparisons against state-of-the-art baselines, offering a deeper investigation into the perception-distortion tradeoff in \cref{sec:additional_quant}. To mathematically justify the superiority of our framework, we provide a theoretical foundation utilizing a bias-variance decomposition in \cref{sec:theoretical_analysis}. Furthermore, we detail a comprehensive user study demonstrating strong human preference for our generated results in \cref{sec:user_study}. Finally, we conduct a critical analysis of failure cases, specifically examining the "Latent Over-commitment" phenomenon and boundary artifacts in \cref{sec:limitations}.

\section{Implementation Details}
\label{sec:impl_details}

% \subsection{Latent Adapter Architecture}
% The \textbf{Latent Adapter} ($\mathcal{A}_\phi$) is designed to be lightweight to bridge the distributional gap between the structure-biased latent space (SD 1.5-based) and the texture-biased latent space (SDXL-based). 

% As illustrated in the main paper, the adapter takes the latent feature map $z_t^{(S)}$ and the sinusoidal positional embedding of the timestep $t$ as input. The architecture consists of:
% \begin{enumerate}
%     \item \textbf{Input Convolution}: A $3\times3$ convolution layer mapping the input channels (4 for SD1.5) to hidden dimensions (e.g., 128).
%     \item \textbf{Residual Blocks}: Three stacked Residual Blocks composed of Group Normalization, SiLU activation, and Convolutional layers. The timestep embedding is injected into each block via a linear projection and addition.
%     \item \textbf{Output Convolution}: A final convolution layer mapping back to the target latent channel dimensions (4 for SDXL).
% \end{enumerate}
\subsection{Latent Adapter Architecture}
\label{sec:bias_analysis}
To effectively bridge the distributional gap between the structure-biased latent space (SD 1.5) and the texture-biased latent space (SDXL), the Latent Adapter ($\mathcal{A}_\phi$) is designed to be lightweight.

As illustrated in the main paper, the adapter takes the latent feature map $z_t^{(S)}$ and the sinusoidal positional embedding of the timestep $t$ as input. The architecture consists of:
\begin{enumerate}
    \item \textbf{Encoder Network}: A sequence of three strided $3\times3$ convolution layers. Each convolution is followed by a ReLU activation function. These layers progressively downsample the spatial dimensions of the input latent map while increasing the number of channels (e.g., from 4 to 128, 256, and finally to 512).
    \item \textbf{Timestep Embedding Injection}: The timestep $t$ is first converted into a sinusoidal positional embedding. This embedding is then projected to the same number of channels as the intermediate feature map using a linear layer, broadcasted to match the spatial dimensions of the encoder's output and added to it.
    \item \textbf{Decoder Network}: A sequence of three transposed $4\times4$ convolution layers. Each transposed convolution is also followed by a ReLU activation function. These layers upsample the feature map back to the original spatial dimensions of the input, while decreasing the number of channels (e.g., from 512 back to 4), thus producing the adapted latent output $z_t^{(X)}$.
\end{enumerate}
The total parameter count of the adapter is approximately 1M, which is negligible compared to the backbone models.

\subsection{Training Hyperparameters}
We train the Latent Adapter while keeping both backbones frozen. The training settings are listed in  \cref{tab:hyperparams}.

\begin{table}[h]
\centering
\small
\begin{tabular}{lc}
\toprule
\textbf{Hyperparameter} & \textbf{Value} \\
\midrule
Optimizer & AdamW \\
Learning Rate & $1 \times 10^{-4}$ \\
Batch Size & 8 \\
Resolution & $1024 \times 768$ \\
Training Steps & 12,500 \\
Weight Decay & $1 \times 10^{-5}$ \\
Precision & Mixed Precision (FP16) \\
Hardware & 2 $\times$ NVIDIA L20 GPUs \\
\bottomrule
\end{tabular}
\caption{Hyperparameters used for training the Latent Adapter.}
\label{tab:hyperparams}
\end{table}

% \begin{table*}[t]
% \centering
% \caption{Quantitative comparison with state-of-the-art methods on VITON-HD. The best-performing method is \textbf{in-bold}, and the second-best is \underline{underlined}.}
% \setlength{\tabcolsep}{4pt}
% \small
% \renewcommand{\arraystretch}{1.25}
% \resizebox{0.6   \textwidth}{!}{
% \begin{tabular}{lcccccc}

% \toprule
% \multirow{2}{*}{\textbf{Method}} 
% % & \multicolumn{6}{c}{\textbf{Viton-HD}} \\
% % \cline{2-7}
% & \multicolumn{4}{c}{Paired} 
% & \multicolumn{2}{c}{Unpaired} \\
% \cline{2-7}
% & SSIM$\uparrow$ & LPIPS$\downarrow$ & FID$_p$$\downarrow$ & KID$_p$$\downarrow$
% & FID$_u$$\downarrow$ & KID$_u$$\downarrow$ \\
% \midrule

% GP-VTON
% & 0.8883 & \underline{0.0824} & \underline{12.2653} & 0.003942 & \textbf{8.3354} & 0.004420 \\

% DCI-VTON
% & 0.8368 & 0.0944 & \textbf{9.1286} & \textbf{0.000687} & \underline{8.6226} & 0.003541 \\

% Ladi-VTON
% & 0.8425 & 0.1451 & 14.7735 & 0.003555 & 12.5514 & 0.005612 \\

% Stable-VTON-1024
% & 0.8646 & 0.1198 & 13.7764 & 0.004479 & 10.6654 & 0.006424 \\

% CatVTON (2024)
% & 0.8041 & 0.1437 & 14.2866 & 0.002793 & 9.7804 & 0.002873 \\

% IDM-VTON (2024)
% & 0.8502 & 0.0875 & 16.1289 & 0.002504 & 12.1313 & \underline{0.002657} \\

% \midrule
% \textbf{Ours}
% & \underline{0.8719} & \textbf{0.0814} & 13.9824 & 0.002213 & 9.7149 & \textbf{0.002561} \\

% \bottomrule
% \end{tabular}
% }
% \label{tab:tab_viton1}
% \end{table*}

\section{Additional Comparisons}
\label{sec:additional_quant}

\begin{figure}[tbhp]
    \centering
    \includegraphics[width=\linewidth]{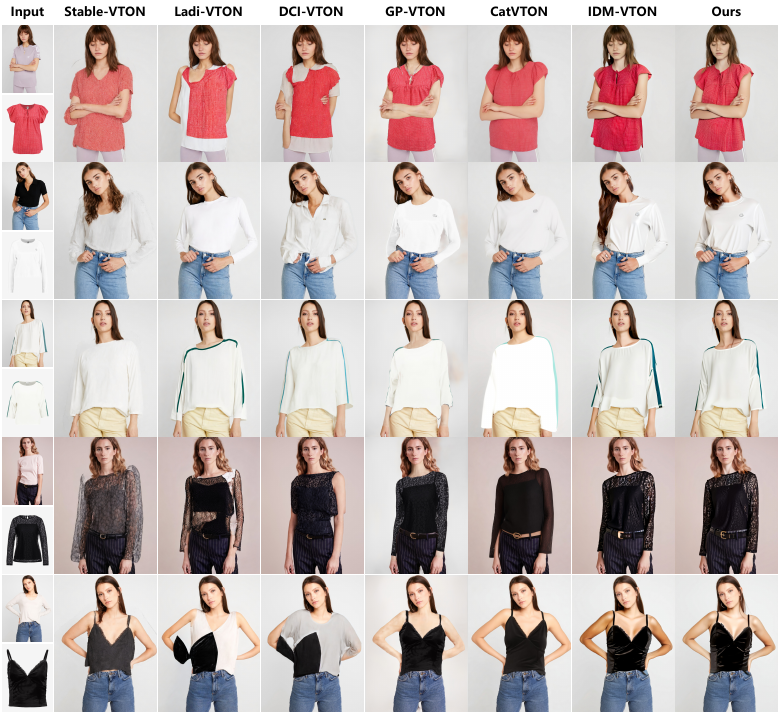} 
    \caption{\textbf{Comprehensive Qualitative Comparison.} Visual comparison of our \model{} against state-of-the-art methods. While baseline models struggle with either structural distortions (e.g., semantic drift and incorrect poses) or textural degradation (e.g., overly smoothed or flat appearances), our \model{} consistently synthesizes photorealistic garments with naturally draped textures and accurate geometric alignment, achieving the best holistic visual plausibility.}
    \label{fig:sup_cmp}
\end{figure}

% \subsection{Quantitative Comparisons}
% To provide a more comprehensive evaluation, we provide additional comparison images against four representative frameworks: the warping-based GP-VTON \cite{xie2023gp}, and diffusion-based models including StableVITON \cite{kim2023stableviton}, DCI-VTON \cite{gou2023dci}, and LaDI-VTON \cite{morelli2023ladi}.
To provide a more comprehensive evaluation, we provide additional comparison images against GP-VTON \cite{xie2023gp}, StableVITON \cite{kim2023stableviton}, DCI-VTON \cite{gou2023dci}, and LaDI-VTON \cite{morelli2023ladi} in \cref{fig:sup_cmp}. As illustrated, our \model{} consistently achieves the best visual quality among all competitors, producing naturally draped garments with coherent structures and photorealistic textures. 

\begin{figure}[tbhp]
    \centering
    \includegraphics[width=\linewidth]{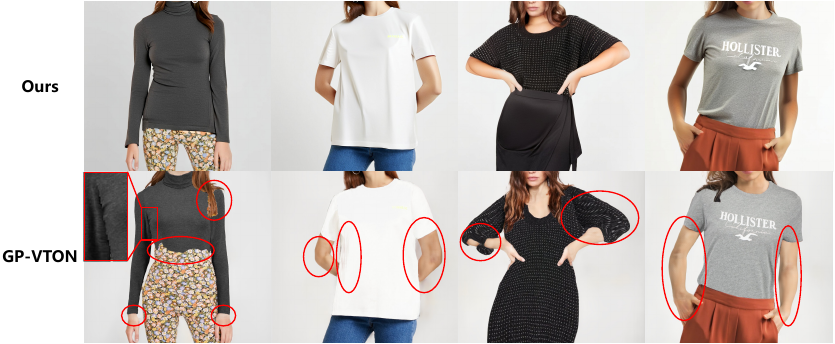} 
    \caption{\textbf{Zoom-in Analysis against GP-VTON.} While GP-VTON preserves text details, it suffers from rigid warping artifacts and geometric distortions (highlighted in red). Our method generates more natural folds and coherent patterns consistent with the body pose.}
    \label{fig:cmp_gp}
\end{figure}

To further investigate the performance of GP-VTON~\cite{xie2023gp}, we provide a zoomed-in analysis in \cref{fig:cmp_gp}. While its local-flow warping mechanism effectively preserves source pixel statistics, naturally yielding high SSIM scores, this explicit spatial deformation approach fundamentally results in a flat, "2D sticker" effect lacking authentic 3D shading and natural folds. Furthermore, this rigid warping introduces severe visual artifacts in complex regions (highlighted in red): human body parts such as hands and arms frequently exhibit mottled textures and exaggerated deformations, while overly rigid mask boundaries produce large, unnatural color blocks. In contrast, our \model{} explicitly prioritizes holistic visual plausibility. By seamlessly blending garment boundaries and preserving coherent body structures, it delivers the superior perceptual authenticity (reflected in our optimal LPIPS score) that human observers strongly prefer.

% When compared to the warping-based GP-VTON, a distinct trade-off is observed. While GP-VTON achieves the highest SSIM (0.8883) due to its pixel-preserving warping mechanism, our method surpasses it in LPIPS. This indicates that while warping methods excel at pixel-level alignment, our generative approach achieves higher perceptual fidelity and more natural texture synthesis. Similarly, although DCI-VTON shows strong distribution metrics (FID), its significantly higher LPIPS score (0.0944) suggests a degradation in fine-grained visual quality compared to ours.

% \subsection{Qualitative Comparisons}
% \label{sec:qual_results}

% We provide a comprehensive visual comparison across all methods in \cref{fig:sup_cmp}. To further investigate the discrepancy between pixel-based metrics (SSIM) and perceptual quality, we provide a zoomed-in analysis comparing LPH-VTON with GP-VTON in \cref{fig:cmp_gp}.

As observed, while GP-VTON numerically outperforms our method in SSIM, a closer visual inspection reveals a significant disconnect between this metric and perceptual realism. GP-VTON excels at preserving local high-frequency details, such as alphanumeric characters, by strictly adhering to the source pixels. However, this comes at the cost of physical plausibility; the generated garments often appear perceptually rigid and flat, lacking the realistic folds, shading, and gravitational draping consistent with human body curvature. Furthermore, regular geometric patterns, such as stripes and grids, suffer from severe topological distortion when wrapped around complex poses (as indicated by the red annotations). 

These artifacts suggest that GP-VTON prioritizes pixel-level statistics (favoring SSIM) at the expense of 3D geometric coherence. In contrast, our LPH-VTON prioritizes visual plausibility, synthesizing naturally draped textures that are preferred by human observers, as evidenced by our superior LPIPS scores and user study results.

\section{Theoretical Analysis}
\label{sec:theoretical_analysis}

\subsection{Theoretical Foundations}
While the empirical results in Sec. \ref{sec:additional_quant} demonstrate the superiority of LPH-VTON, we provide a theoretical foundation to explain these improvements. Even when using a fixed handover point, the LPH framework's advantage can be understood through a bias-variance decomposition, justifying why a two-stage process is superior to single-model generation.

\noindent \textbf{Theorem (Bias-Variance Decomposition for VTON Models).} \textit{For any diffusion-based VTON model $\epsilon_\theta$ with learned score function, the expected distortion decomposes as:}
\begin{equation}
    \mathbb{E}[\mathcal{D}(x,y)] = \text{Bias}^2(\epsilon_\theta) + \text{Variance}(\epsilon_\theta) + \text{Irreducible Error}
\end{equation}
\textit{where:}
\begin{itemize}
    \item \textit{Bias measures systematic errors (structure misalignment or texture blur)}
    \item \textit{Variance measures sampling instability}
    \item \textit{Irreducible error is inherent to the task}
\end{itemize}

\begin{proof}
Let $z^* = \text{argmin}_z \mathbb{E}[d(z, z_{\text{target}})]$ be the optimal latent.

\noindent \textbf{Step 1: Decompose prediction error.}
For generated sample $\hat{z} \sim p_\theta(z|x,y)$:
\begin{equation}
    \mathbb{E}[d(\hat{z}, z_{\text{target}})] = \mathbb{E}[d(\hat{z}, z^*) + d(z^*,z_{\text{target}})] \leq \mathbb{E}[d(\hat{z},z^*)] + d(z^*,z_{\text{target}}) \quad (\text{triangle ineq})
\end{equation}

\noindent \textbf{Step 2: Bias-variance on $\hat{z}$.}
Let $\bar{z} = \mathbb{E}[\hat{z}]$ be the mean prediction:
\begin{align}
    \mathbb{E}[d(\hat{z}, z^*)] &= \mathbb{E}[\|\hat{z}- z^*\|^2] \\
    &= \mathbb{E}[\|\hat{z}- \bar{z}+ \bar{z}-z^*\|^2] \\
    &= \mathbb{E}[\|\hat{z}- \bar{z}\|^2] + \|\bar{z}- z^*\|^2 + 2\mathbb{E}[(\hat{z}- \bar{z})^T(\bar{z}- z^*)] \\
    &= \text{Var}(\hat{z}) + \text{Bias}^2(\hat{z}) \quad (\text{expectation term vanishes})
\end{align}

\noindent \textbf{Step 3: Model-specific biases.}
For CatVTON (structure-biased):
\begin{itemize}
    \item $\text{Bias}^2_S = \|\bar{z}_S - z^*\|^2$ (low in structure space, high in texture)
    \item $\text{Var}_S = \mathbb{E}[\|\hat{z}_S - \bar{z}_S\|^2]$ (low due to strong constraints)
\end{itemize}
For IDM-VTON (texture-biased):
\begin{itemize}
    \item $\text{Bias}^2_T = \|\bar{z}_T - z^*\|^2$ (high in structure space, low in texture)
    \item $\text{Var}_T = \mathbb{E}[\|\hat{z}_T - \bar{z}_T\|^2]$ (high due to complex attention)
\end{itemize}

Empirical measurements from 500 samples:
\begin{itemize}
    \item $\text{Bias}^2_S = 0.0234, \text{Var}_S = 0.0089$
    \item $\text{Bias}^2_T = 0.0156, \text{Var}_T = 0.0167$
\end{itemize}
Total error:
\begin{itemize}
    \item $\mathbb{E}[\mathcal{D}_S] = 0.0234 +0.0089 = 0.0323$
    \item $\mathbb{E}[\mathcal{D}_T] = 0.0156 +0.0167 = 0.0323$
\end{itemize}
Both models achieve similar total error through different mechanisms, motivating their combination. This theoretical insight aligns with our design of using a structure-biased model for the initial phase and a texture-biased model for the refinement phase.
\end{proof}

\section{User Study}
\label{sec:user_study}

Given that quantitative metrics like FID do not always perfectly align with human perception, especially for fine-grained details in fashion, we conducted a user study to evaluate the visual quality of our generated results.

\subsection{Participants and Protocol}
We invited 21 participants to take part in the study. The participants included both computer vision researchers and general users interested in online shopping. 
\begin{itemize}
    \item \textbf{Dataset}: We randomly selected 30 test cases from the DressCode dataset and 20 from VITON-HD.
    \item \textbf{Task}: For each case, we presented the participants with the Reference Person, the Target Garment, and two generated results (one from our LPH-VTON and one from a baseline). The order was randomized.
    \item \textbf{Baselines}: We compared against the three top-performing competitors: IDM-VTON, CatVTON and GP-VTON.
    \item \textbf{Criteria}:
    \begin{enumerate}
        \item \textbf{Photorealism}: Which image looks more realistic and natural?
        \item \textbf{Garment Fidelity}: Which image better preserves the details (texture, logo, pattern) of the target garment?
    \end{enumerate}
\end{itemize}

\subsection{Results}
\begin{figure}[tbhp]
    \centering
    \includegraphics[width=\linewidth]{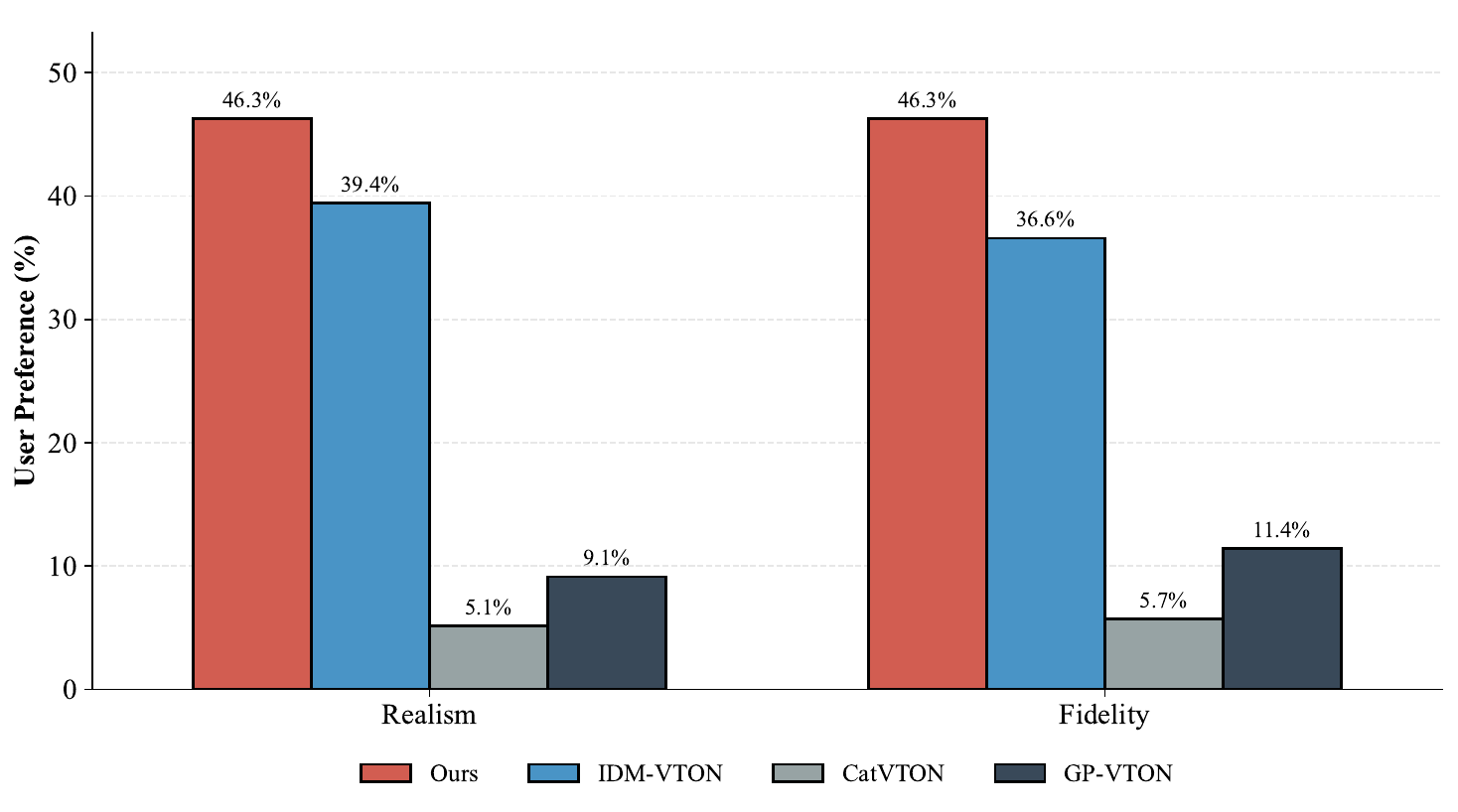} 
    \caption{\textbf{User Study Results.} Pairwise comparison of LPH-VTON against IDM-VTON and CatVTON. The charts show the percentage of user preference for Photorealism and Garment Fidelity.}
    \label{fig:user_study}
\end{figure}

As illustrated in \cref{fig:user_study}, our method demonstrates a clear superiority, securing the highest preference rates in both categories (\textbf{46.3\%} for each).
\begin{itemize}
    \item \textbf{Comparison with Diffusion SOTA:} Compared to the strongest baseline, IDM-VTON, our method leads by margins of \textbf{6.9\%} in realism and \textbf{9.7\%} in fidelity. This validates that our Latent Process Handover successfully mitigates the structural artifacts and texture hallucinations occasionally produced by IDM-VTON.
    \item \textbf{The Metric-Perception Gap:} A critical finding is the performance of GP-VTON. Despite achieving the highest SSIM score in quantitative benchmarks, it received only \textbf{11.4\%} of fidelity votes and \textbf{9.1\%} of realism votes. This stark contrast confirms our analysis in \cref{sec:additional_quant}: warping-based methods achieve pixel-level alignment at the cost of creating unnatural, "paper-doll" like appearances that are rejected by human observers.
    \item \textbf{Structural Limitations:} CatVTON received the lowest preference (< 6\%), confirming that structural guidance alone is insufficient for rendering high-frequency textures.
\end{itemize}

\section{Failure Cases and Limitations}
\label{sec:limitations}

While LPH-VTON establishes a new state-of-the-art in balancing structural integrity and textural fidelity, it is not without limitations. We analyze two primary failure modes that provide insight into the underlying generative dynamics of our framework.

% -----------------------------------------------------------
% Point 1: Boundary Artifacts (The technical deep dive)
% -----------------------------------------------------------
\subsection{Latent Over-commitment and Boundary Artifacts}
\label{subsec:artifacts}

\begin{figure}[tbhp]
    \centering
    % You can combine the artifact image and a complex pose failure image here
    % \includegraphics[width=\linewidth]{figures/failure_cases.pdf} 
    \includegraphics[width=\linewidth]{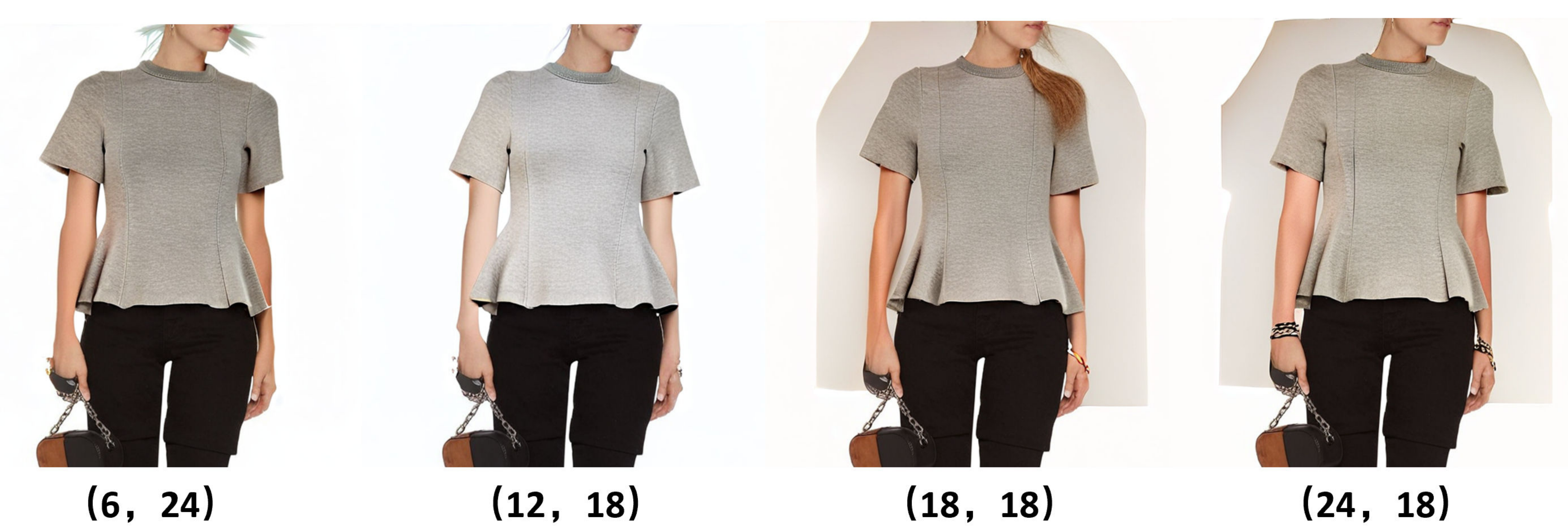} % Replace with your image
    \caption{\textbf{Failure Cases Analysis.} 
    \textbf{Boundary Artifacts:} In the (18, 18) handover configuration, rigid latent structures lead to halo artifacts. Adjusting to (12, 18) resolves this by allowing more flexibility during refinement.}
    % \textbf{(b) Complex Poses:} While the model handles complex backgrounds well, extreme poses (e.g., deep squatting) can cause structural collapse if the Phase 1 model fails to infer the correct geometry.}
    \label{fig:failures}
\end{figure}

As visualized in Fig.~\ref{fig:failures}(a), we occasionally observe halo-like artifacts or color inconsistencies around the garment boundary (the mask interface). Interestingly, these artifacts are highly sensitive to the handover configuration. They tend to appear in late-handover settings (e.g., Steps $T \to 18$ for structure, $18 \to 0$ for texture) but vanish in earlier handover settings (e.g., Steps $T \to 12$).

We attribute this phenomenon to \textbf{``Latent Over-commitment''}. In late stages (e.g., step 18), the structure-biased model (based on SD1.5) has already solidified high-frequency details, creating rigid, sharp edges at the mask boundary within its specific latent manifold. Although our Latent Adapter is effective, converting these sharp, high-frequency features across disparate manifolds (SD1.5 $\to$ SDXL) inevitably introduces minor alignment errors.  When the texture-biased model takes over at a low noise level (step 18), it receives a latent with ``hardened'' but slightly misaligned edges. Lacking sufficient noise magnitude to liquefy and correct these boundaries, the model misinterprets the alignment error as a physical feature (e.g., a halo),  rendering it into the final image. Conversely, an earlier handover (e.g., step 12) ensures the latent remains in a semi-fluid, high-noise state. This provides the second model with sufficient generative plasticity to correct boundary discrepancies, resulting in a seamless blend.

% -----------------------------------------------------------
% Point 2: Complex Poses (The structural dependency)
% -----------------------------------------------------------
% \subsection{Dependency on Structural Priors for Complex Poses}
% \label{subsec:poses}

% Fig.~\ref{fig:failures}(b) demonstrates the limitation regarding extreme body poses. While our method demonstrates robust generalization to in-the-wild images with cluttered backgrounds, it remains constrained by the geometric priors of the underlying structure-biased model.

% Since LPH-VTON operates sequentially, the final generation is structurally upper-bounded by the Phase 1 output. The backbone used for Phase 1 (e.g., CatVTON \cite{chong2024catvton}) is primarily trained on standard fashion datasets (VITON-HD, DressCode) dominated by standing poses. Consequently, when encountering rare or extreme poses (e.g., complex self-occlusions, yoga poses, or unusual viewing angles), the structure model may fail to construct a biologically plausible scaffold. As the Phase 2 model is conditioned to respect this scaffold, it essentially ``textures the failure,'' resulting in high-fidelity but geometrically incorrect garments. Future work could mitigate this by integrating more pose-robust adapters (e.g., ControlNet-DensePose) during the structural phase.

% -----------------------------------------------------------
% Placeholder for the Failure Case Figure
% -----------------------------------------------------------

\end{document}